\documentclass[conference]{IEEEtran}
\IEEEoverridecommandlockouts

\usepackage{cite}
\usepackage{amsmath,amssymb,amsfonts}
\usepackage{algorithmic}
\usepackage{graphicx}
\usepackage{textcomp}
\usepackage{xcolor}
\usepackage{url}

\usepackage[a4paper]{geometry}
\def\BibTeX{{\rm B\kern-.05em{\sc i\kern-.025em b}\kern-.08em
    T\kern-.1667em\lower.7ex\hbox{E}\kern-.125emX}}
\begin{document}

\title{Collaborative real-time vision-based device for olive oil production monitoring}

\author{
    \IEEEauthorblockN{1\textsuperscript{st} Matija Šuković}
    \IEEEauthorblockA{\textit{Faculty of Natural Sciences and Mathematics} \\
    \textit{University of Montenegro}\\ Cetinjska 2, 81000 Podgorica, Montenegro
    \\
    matija.sukovic23@gmail.com}
    \and
    \IEEEauthorblockN{2\textsuperscript{nd} Igor Jovančević}
    \IEEEauthorblockA{\textit{Faculty of Natural Sciences and Mathematics} \\
    \textit{University of Montenegro}\\
    Cetinjska 2, 81000 Podgorica, Montenegro \\
    igorj@ucg.ac.me}
}

\maketitle

\begin{abstract}
This paper proposes an innovative approach to improving quality control of olive oil manufacturing and preventing damage to the machinery caused by foreign objects. We developed a computer-vision-based system that monitors the input of an olive grinder and promptly alerts operators if a foreign object is detected, indicating it by using guided lasers, audio, and visual cues.
\end{abstract}


\vspace{10pt}
\noindent{} © 2024 IEEE.  Personal use of this material is permitted.  Permission from IEEE must be obtained for all other uses, in any current or future media, including reprinting/republishing this material for advertising or promotional purposes, creating new collective works, for resale or redistribution to servers or lists, or reuse of any copyrighted component of this work in other works.

\section{Introduction}

\begin{figure}[htbp]
    \centerline{
        \includegraphics[width=150px]{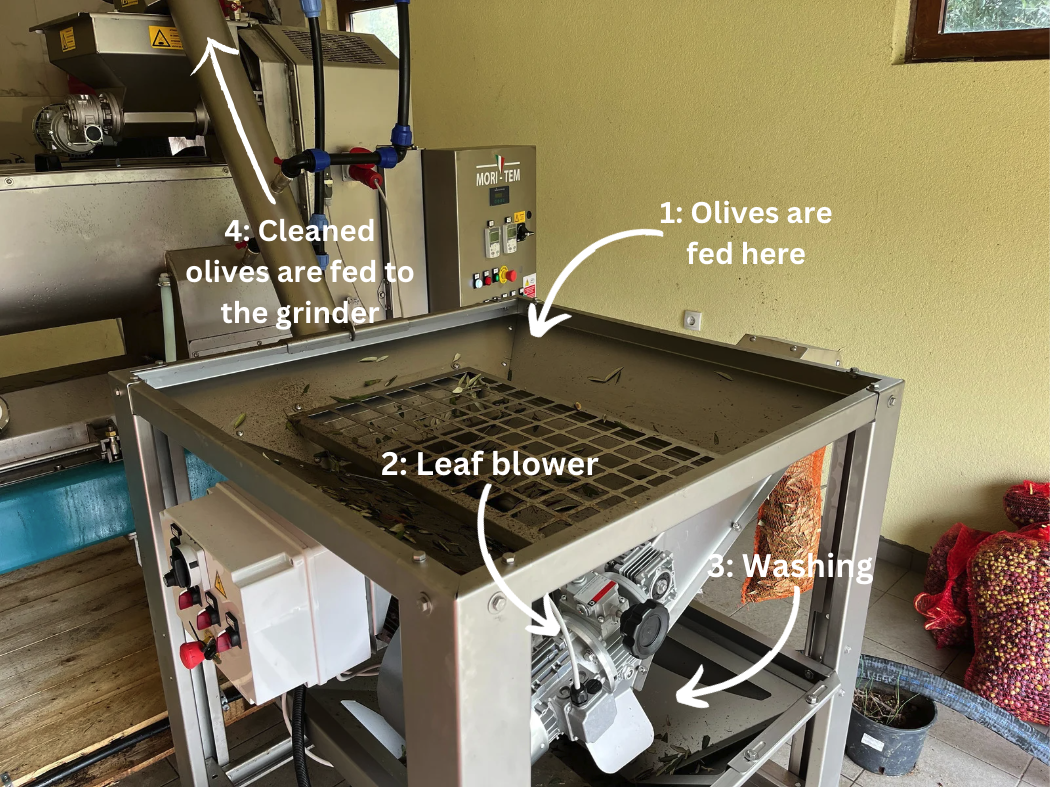}
    }
    \caption{A washing unit used to automatize the preprocessing of the olives}
    \label{washing_unit}
\end{figure}

An olive grinding machine is used to extract extra virgin olive oil from freshly harvested olives. An entry-level olive grinder is a small unit targeted for medium to large-sized olive farms. Grinders like these require the fruits to be washed and cleaned from any foreign objects, such as leaves or branches, before entering the machine. To reduce the manual labor of cleaning the olives, these grinders are commonly paired with a washing unit, such as the one shown in Fig.~\ref{washing_unit}, which automatically washes the olives and removes leaves and small branches. 

While it decreases the manual labor, the washing unit indirectly increases the risk of foreign objects, e.g. rocks or small tools and pieces, entering the grinder, due to the reduced manual inspection of olives. Rocks are especially problematic, as they are often picked together with the fruit since olive trees thrive on rocky terrain. They can have similar shape, size, and even color to the fruit, making them hard to spot. Yet if one enters the grinder, the machinery needs to be halted right away as the rock will wear down the blades at best, and jam the machine causing immense damage at worst. In such case, the grinder has to be halted and opened to extract the foreign object(s), which exposes the crushed olives to air, making them unsuitable for extra virgin olive oil and causing them to be discarded.

The washing unit is designed in such a way that it allows for an easy visual inspection as the olives are dumped into it, by having a wide input area for the fruits to spread as they are slowly fed to the machine. The operating personnel is expected to perform a manual inspection as they dump freshly harvested olives from bags and boxes. We propose to enrich this process with a device that will perform an automatic visual inspection powered by computer vision, reducing strain on workers and decreasing the risk of rocks entering the grinder, thus lessening the chance of expensive damage happening to the machinery and, in general, increasing the quality of the end product.

\section{Solution concept}

\begin{figure}[htbp]
    \centerline{
        \includegraphics[width=100px]{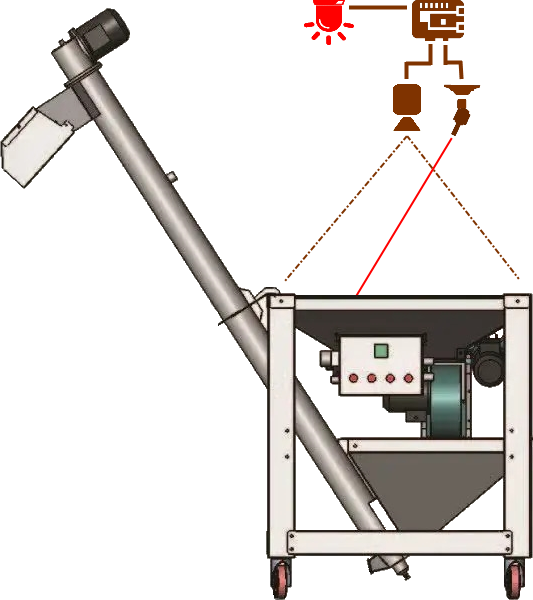}
    }
    \caption{Concept of the proposed solution}
    \label{concept_image}
\end{figure}

The concept of the prototype solution, shown in Fig.~\ref{concept_image}, can be thought of as a digital assistant in the olive oil production process. The device is to be installed above the washing unit. It features LED lights and a camera facing down, overlooking the entrance of the machine. The frames captured by the camera are fed to a single-board computer running an Artificial Intelligence (AI) model that detects the rocks. If any are detected, the device fires a series of audio and visual alerts to draw the workers' attention. Most notably, it guides a laser pointer to a detected rock by controlling a pan-tilt head with a laser diode. The laser beam makes it trivial for the workers to locate and extract the rock before it enters the machine.

The target hardware to run the AI model on is either a Raspberry Pi or a device from the NVIDIA Jetson series, depending on how demanding the final model will be. The goal is for the end product to be low-cost and thus affordable by small olive farms that use entry-level grinders. With this in mind, we will focus on tuning and optimizing the model to run well on the cheaper Raspberry Pi, resorting to a more expensive Jetson device if necessary.

For the camera of the prototype, we chose a Raspberry Pi Camera Module v3. This camera is suitable for our application for several reasons. It has a high-quality sensor capable of capturing images at a resolution of 12MP (megapixel). It also has auto-focus built-in, which helps in keeping the image sharp no matter how many olives are contained in the washing unit. Lastly, it is compact enough to not get in the way of workers, and it is designed to be easily integrated with embedded devices. For future prototypes we will consider using a 3D camera in order to improve the laser tracking, or fitting the camera to a pan-tilt motor to give the device added functionalities, such as surveillance of the machinery while the washing unit is not operating.

\section{Related work}

The problem we are tackling in this work is fairly unique in its nature. Foreign object detection in agriculture and food industries is done in controlled environments designed to provide the best possible visibility and clarity of the scene, whereas we are building on top of an existing system which was not optimised for automatic inspection. 

A solution for a similar use case was proposed by a team from the ICT Research Institute in Korea, where they used deep learning methods to detect foreign objects in almond and green onion flake food processing \cite{almonds}. In their work, the authors present a synthetic method of obtaining a dataset for visual food inspection, where highly spread-out objects are exposed to several light sources, including a light platform which makes objects easily detectable on the bright background. They also trained a deep learning model to perform image segmentation on datasets acquired with the proposed method.

The lab conditions created by the authors may be achievable in highly-automated food processing, but for our use case we need to rely on the surrounding conditions as little as possible, since our device will not be working in a controlled environment. The task of our model is much more complex because of this, and so we need real-world data as opposed to creating synthetic datasets. The authors haven't taken model optimisation into account since it is not a concern for their use case, but in our work we will need to optimise the model as much as possible, as it is meant to run on low-powered embedded devices. As opposed to image segmentation done by the authors, where objects are detected on pixel-level, our model will be an object detector, which is only concerned with finding the bounding box of the object. This information is enough to point a laser to the center of the bounding box, and thus the detected object.

\section{Data acquisition}

\begin{figure}[htbp]
    \centerline{
        \includegraphics[width=175px]{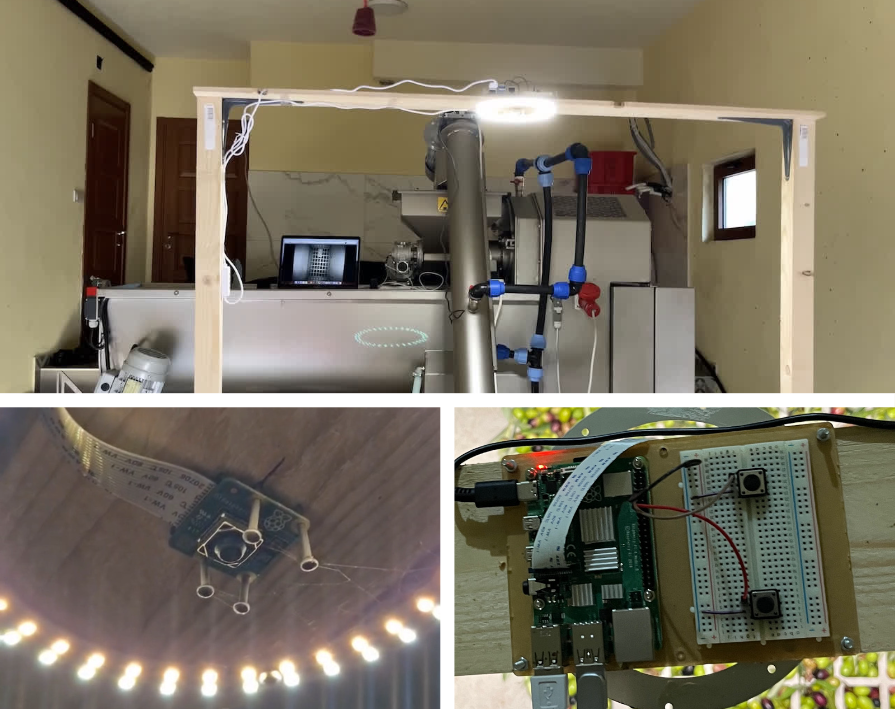}
    }
    \caption{Setup used to acquire the dataset}
    \label{data_acquisition}
\end{figure}

To develop a supervised AI model capable of detecting foreign objects, we were required to collect data of sufficient quality and amount. For this purpose, we obtained access to an olive grinder during the harvesting season, where we could set up equipment for data acquisition during normal operation of the machine (Fig.~\ref{data_acquisition}). A wood construction was built to hold a camera above the input of the washing unit. A ring of LED lights was installed around the camera lens, which helped make the foreign objects more apparent and reduced the impact of the day-night cycle to the quality of the data. These lights also proved to be useful to human operators during the manual inspection process. The camera was connected to a single-board computer, which would start taking pictures at a rate of one per second whenever the washing unit was operating.

The equipment was installed at the beginning of the harvesting season in late October, and it was collecting data until the end of the season around mid December. In this period, 81\,304 images were obtained in total. In order to obtain enough positive samples, foreign objects were often manually tossed in the machine whenever it would be safe to do so. During the first month, these objects were a variety of rocks, debris and tools. It became evident, however, that rocks are the most common naturally occurring foreign object, and so for the remainder of the harvesting season we shifted our focus to acquiring as many samples of rocks as possible.

\begin{figure}[htbp]
    \centerline{
        \includegraphics[width=165px]{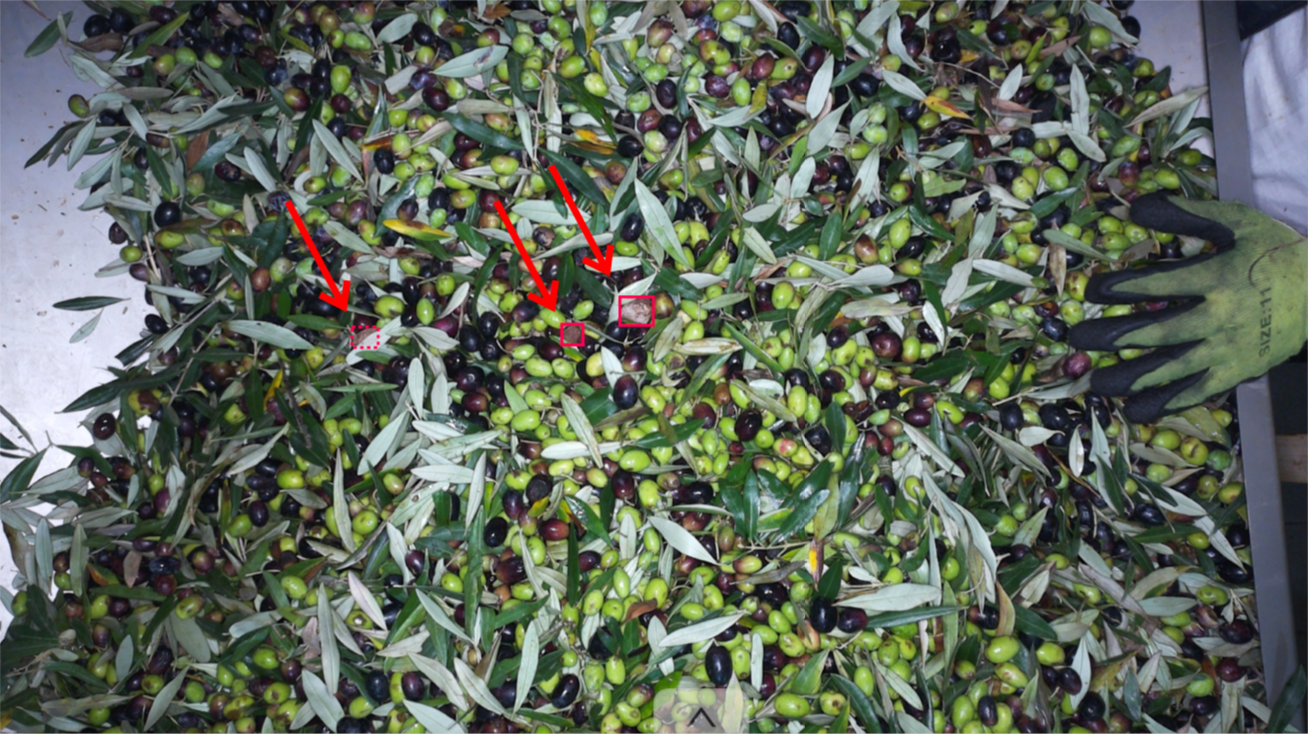}
    }
    \caption{Example dataset image. Three rocks can be seen here}
    \label{dataset_example}
\end{figure}

The data acquisition process resulted in two substantial datasets: one for general foreign object detection in olive oil production, and one focusing on rock detection for the same use case. The latter was cleaned and annotated, resulting in a dataset counting 1878 images containing 5245 annotations. An example image of this dataset is shown in Fig.~\ref{dataset_example}.

Due to the high frequency of picture taking, the raw dataset contains a large amount of highly correlated images. The first step in cleaning the dataset was to reduce the amount of similar images, as these would not be beneficial for model training, and could even cause overfitting to the train data. A fingerprinting algorithm called 'dHash'\cite{dHash} was applied to the images. This algorithm generates hash codes from images by crushing them down to a size of 9x8, applying grayscale transformation, and using the differences between adjacent pixels as the input to a hash function. The resulting hashes will be similar to each other if the respective source images are similar. A similarity threshold of 98\% was used when grouping similar images. For each group, one or more images were manually selected for preservation depending on the quality of their content, before deleting the rest.

During the annotation process, further cleaning was applied by manually deleting some images. For example, if a particular rock appeared in too many consecutive frames without moving or changing position, some of those frames would be discarded to prevent overfitting.

\begin{figure}[htbp]
    \centerline{
        \includegraphics[width=165px]{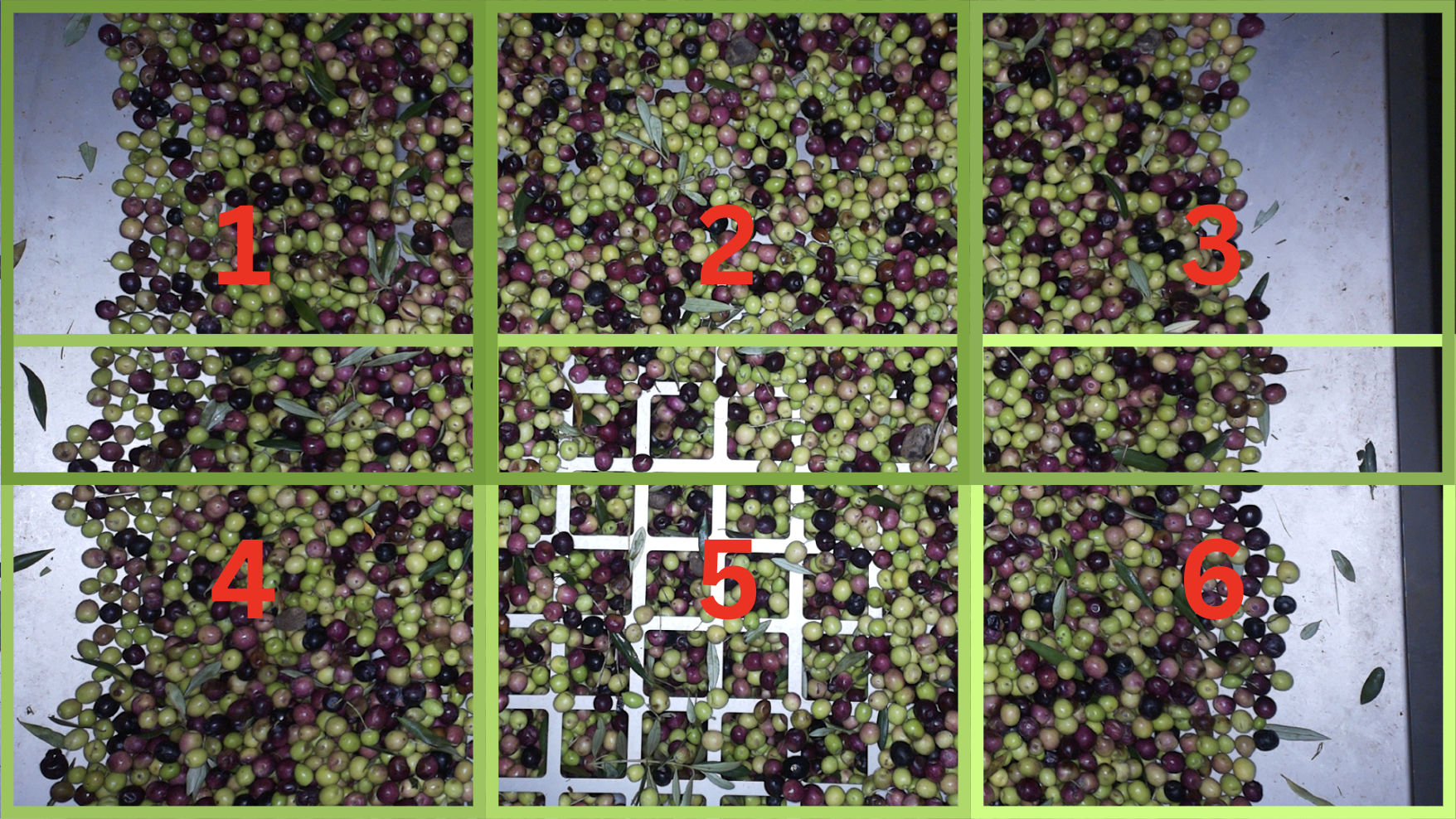}
    }
    \caption{Tiling a source image into six smaller patches}
    \label{tiling}
\end{figure}

We applied tiling to the training dataset to improve our model's performance on detecting small objects. Every image in the dataset was split into patches sized 640x640 each (Fig.~\ref{tiling}). This size was chosen because it is commonly used in pre-trained models, and is thus the best image size to use when fine-tuning these models. Our original images are sized 1920x1080, and tiling produced six patches for each image. The patches overlap on the $y$ axis by roughly 37\%, but there is no overlap on the x axis due to the width of the original image being divisible by the width of the patch. This is not ideal, since having some overlap is useful to prevent loss of data in cases where an annotation is located at the line on which the image splits. This is a rare occurrence in our case, since our target objects are so small, it is unlikely that they will be split by tiling. The overlap on the $y$ axis should ideally be less then what we have. As we work on training new models, we will experiment with various patch sizes to see what gives us the best results. We will need to take into account the inference time, as increasing the number of patches will have a negative effect on the time it takes the model to process a single image. At inference-time, we are using SAHI (Slicing Aided Hyper Inference)\cite{SAHI} to tile the input image, run inference on each patch, and combine the results into the final predictions for the original image. 

\begin{figure}[htbp]
    \centerline{
        \includegraphics[width=165px]{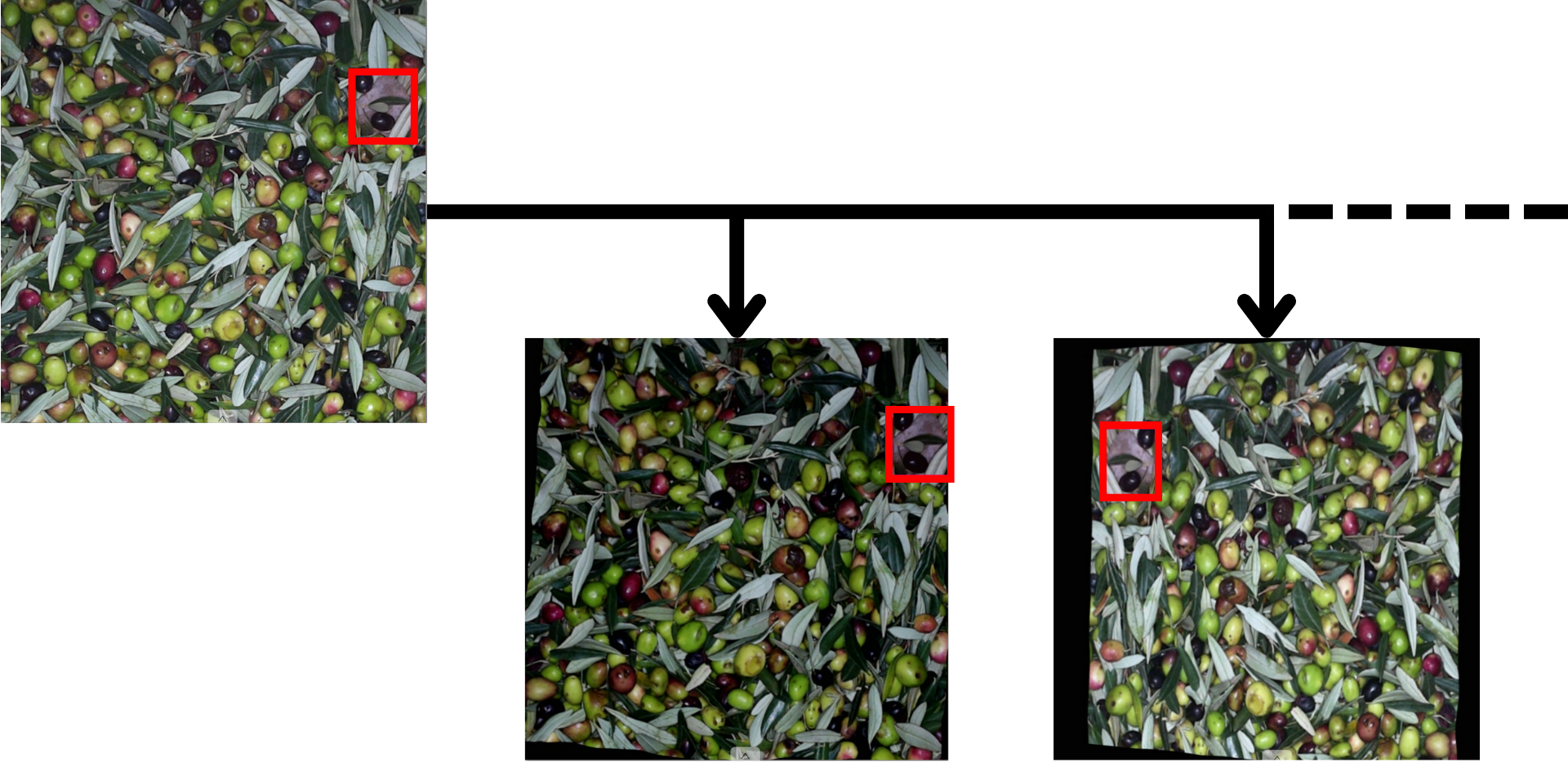}
    }
    \caption{The augmentation pipeline can produce as many augmented samples as it is needed}
    \label{augmentations}
\end{figure}

In order to provide our model more examples of target objects, various transformations were applied to the original dataset. This is a common practice when working with computer vision tasks, where the existing dataset is expanded by applying transformations such as flips, rotations, color tweaks and so on. Using the Albumentations library\cite{albumentations}, a data augmentation pipeline was created through which images were passed. The pipeline contains various augmentations, each having a set probability for being applied. Images can be flipped horizontally and vertically, randomly rotated up to 10 degrees, or randomly rotated by 90 degrees. Some transformations were always applied, to ensure that the pipeline never generates exact duplicates. For example, the pipeline would always apply a random shift in brightness and contrast, as well as CLAHE (Contrast Limited Adaptive Histogram Equalization) \cite{CLAHE}. Lastly, it would apply an elastic transform to the image, which generates a random displacement vector for every pixel. All the mentioned transformations give results diverse enough to allow us to generate randomly augmented samples on demand, by passing the original images through the pipeline as many times as it is required. (Fig.~\ref{augmentations})

\section{Object detection model}

With the dataset ready, work began on training an AI model for rock detection. This proved to be a difficult task for several reasons. The images we are working with (Fig.~\ref{dataset_example}) are high-resolution and contain many small objects. Since the targeted rocks can be so similar to the background olives, our dataset has a very low signal-to-noise ratio, making it difficult for models to extract features that are specific to the target object. One way we combat this problem is by adding an amount of background images (images that contain no target objects) to the dataset, which helps the model better understand the subtle differences between olives and rocks.

\begin{figure}[htbp]
    \centerline{
        \includegraphics[width=125px]{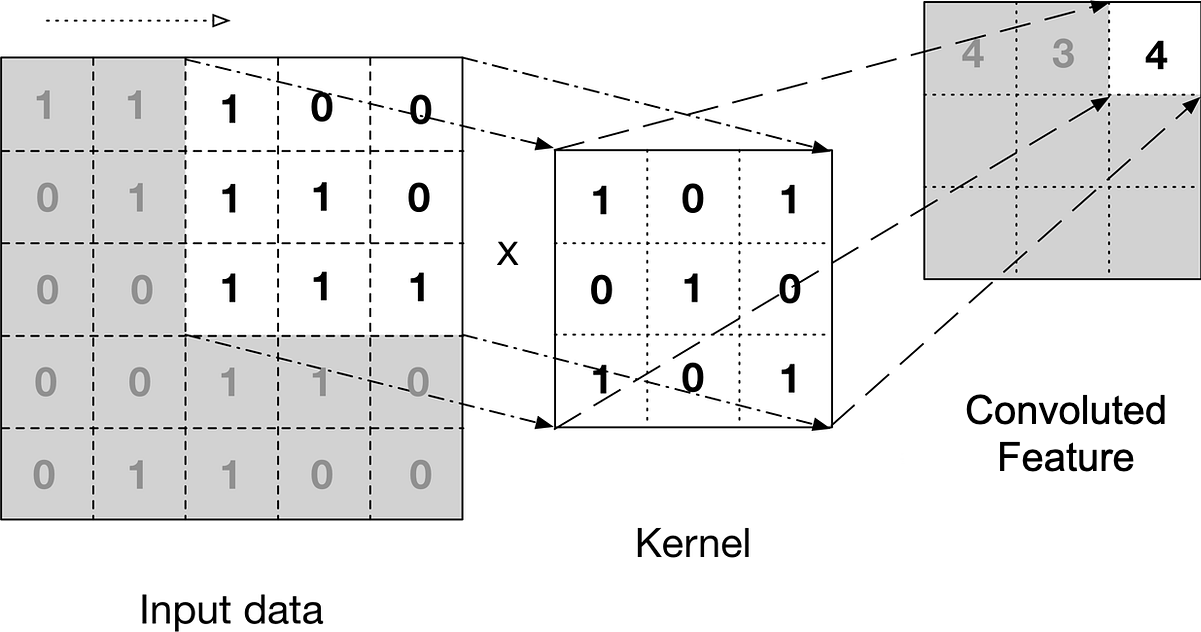}
    }
    \caption{Depiction of a convolutional layer in a CNN}
    \label{convolutional_layer}
\end{figure}

Another issue comes from the fact that our target objects can be very small compared to the image size. This poses a problem for state-of-the-art computer vision models since they are all based on Convolutional Neural Networks (CNN). The main building block of these neural networks is the convolutional layer, where a kernel is passed over an image, performing a convolution operation, in essence summarising several adjacent pixels into a single value (Fig.~\ref{convolutional_layer}). With each passage through a convolutional layer, more abstract features are extracted from the image. The problem with small objects in large images is that their features can get lost after passing the first few convolutional layers of the neural network, being overwhelmed by the features of the background. In order to surpass this issue, on top of tiling the dataset, additional changes needed to be applied to the model architecture. 

For the development of the prototype, we settled with the YOLO version 8\cite{YOLOv8} architecture. This is a state-of-the-art computer vision model which strikes a good balance between complexity and accuracy. When compared to some popular alternatives, it is slightly less accurate while being much faster and less demanding. This is important for our use case, as the trained model will have to run on a low-power embedded device, so it needs to be simple and well optimised. YOLO offers several model types, ranging from 'nano' to 'extra large'. The models get progressively more complex with slower inference time, but higher overall performance. For this prototype we use a modified version of the 'large' model, with an added P2\footnote{P2 refers to the number of strides of the feature pyramids used in the model.} head built specifically to increase performance on detecting smaller objects.

\section{Laser tracking}

In this section, we describe the hardware and software behind the laser tracking feature of the prototype. 

Once the AI model detects a rock amongst the olives, the device needs to communicate that information to the nearby operators as soon as possible in a simple and straightforward manner, so that the threat can be safely handled. Flashing lights and warning sounds can attract the operators' attention, but they are not helpful when it comes to pinpointing the location of the foreign object. Our model, being an object detector, has that information, so the device ought to pass it to the human operators somehow. A display showing the location of the threat would suffice, but this approach generates some friction during the workflow, as operators now need to analyze the display, and then map what they see to the real world. To reduce this friction, we decided to implement a laser tracking system, where the device can use a laser head to point to the rock. This approach ensures that dangerous objects can be removed as quickly as possible, with the least amount of strain on the operator.

Suffice it to say that mounting a controllable laser head to our device would pose a serious threat to the retinas of the surrounding workers. Eyesight damage is a real health hazard when working with lasers, so we needed to ensure the safety of the users of our device. Lasers with a rated power of less than 5mW are deemed safe for the human eye, as the reflex to blink will kick in before the laser can damage the eye's retina. The problem is that these lasers are simply not powerful enough for our purpose, since their beam is not sufficiently visible under the strong lights of our device. We needed to utilize more powerful diodes, but that meant we had to deal with a health hazard. If left as is, workers would be required to wear protective goggles during the operation. This would hinder their ability to perform their tasks, so we never considered it as a viable option. We solved this problem by mounting an adjustment lens to the laser diode. With the added lens, we were able to scatter the laser beam enough that it no longer poses a threat, while still being sufficiently visible. 

\begin{figure}[htbp]
    \centerline{
        \includegraphics[width=125px]{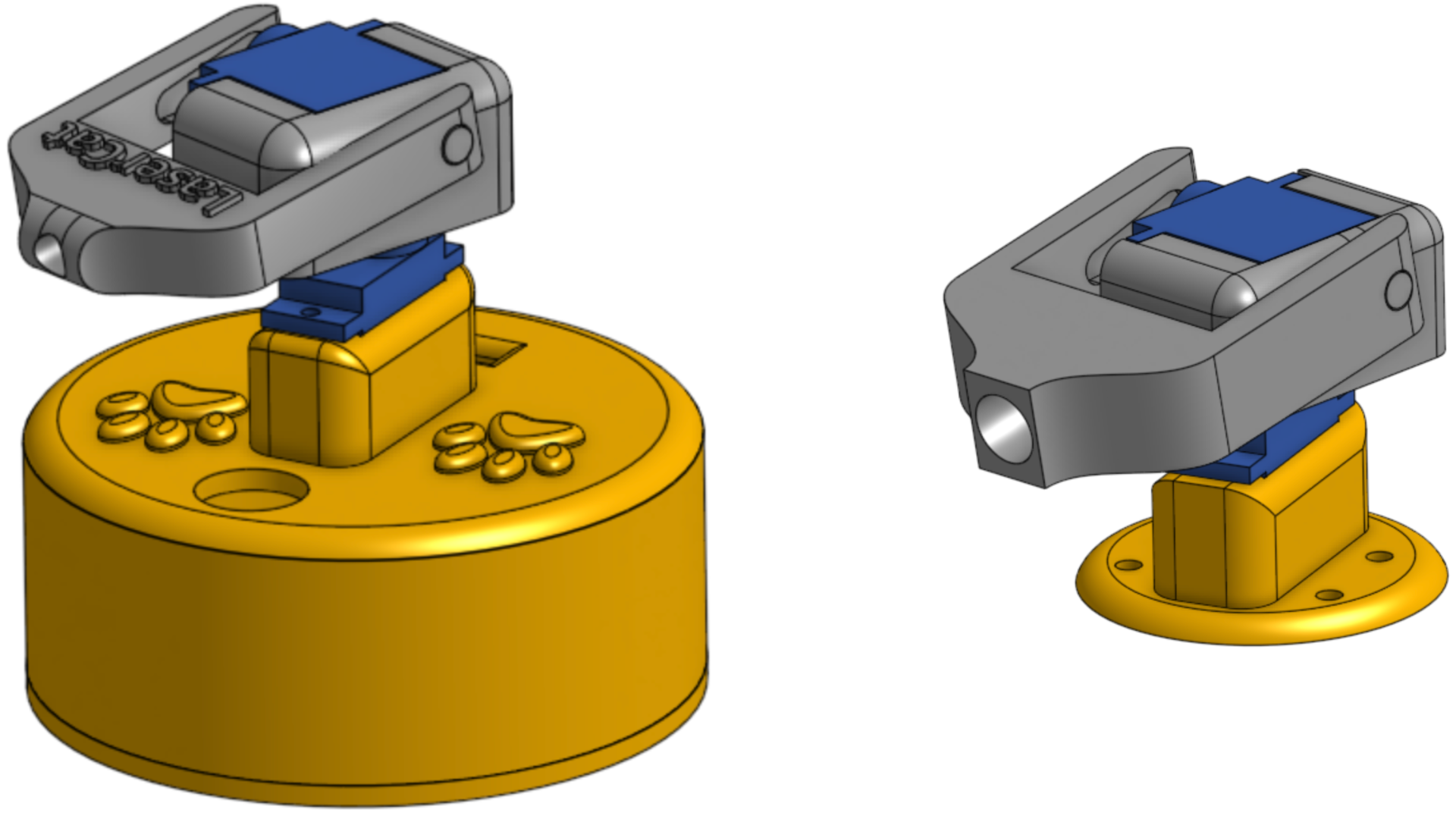}
    }
    \caption{Laser pan-tilt head: Original (left) and modified (right)}
    \label{laser_heads}
\end{figure}

For the chassis of the pan-tilt motor controlling the laser, we modified and repurposed a CAD (Computer Aided Design) model (Fig.~\ref{laser_heads}) from an open source DIY (Do It Yourself) project called LaserCat, featuring a toy designed to entertain cats by shooting a laser around a room for them to chase. The design utilizes two micro servo motors to allow for the pan-tilt motion of a head containing a laser diode. 

\begin{figure}[htbp]
    \centerline{
        \includegraphics[width=135px]{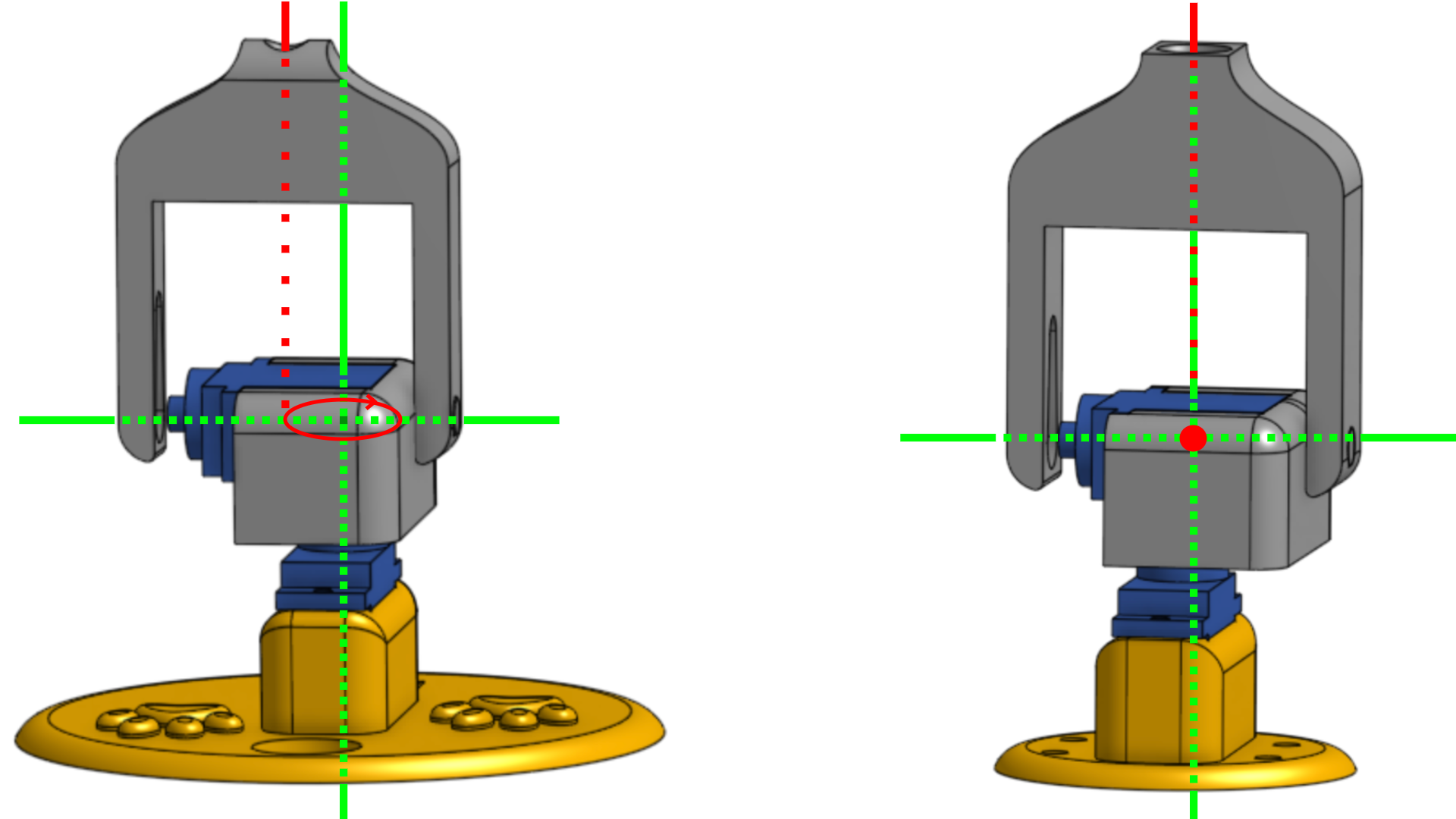}
    }
    \caption{Geometry comparison of the original (left) and modified (right) models}
    \label{heads_comparison}
\end{figure}

Two major modifications were applied to the original CAD model. Firstly, the author designed the laser head with a 5mW laser diode in mind, which can be commonly found in a 6mm form factor. We are using a more powerful diode for our project, so the head had to be modified to be able to house a larger 9mm diode. Secondly, in order to make calculations easier when performing laser tracking, the general geometry of the pan-tilt motor had to be modified. For example, in the original model the position of the laser diode is offset in relation to the rotation axis of the pan servo (Fig.~\ref{heads_comparison}). It would require performing additional calculations to mitigate this and precisely point a laser to a target. Instead of doing that, the CAD model was modified so that the laser diode lays exactly on the pan motor's rotation axis. With this modification, the origin of the laser beam can be mapped to a single point in space, no matter the angle of the two servo motors. 

The laser head was 3D printed and mounted to the same construction that was used for data acquisition. Using this setup we were able to develop and test the laser tracking software, which takes a point on an image captured by the camera and calculates the angles the two servos need to take for the laser to shine on the target point. 

\begin{figure}[htbp]
    \centerline{
        \includegraphics[width=300px]{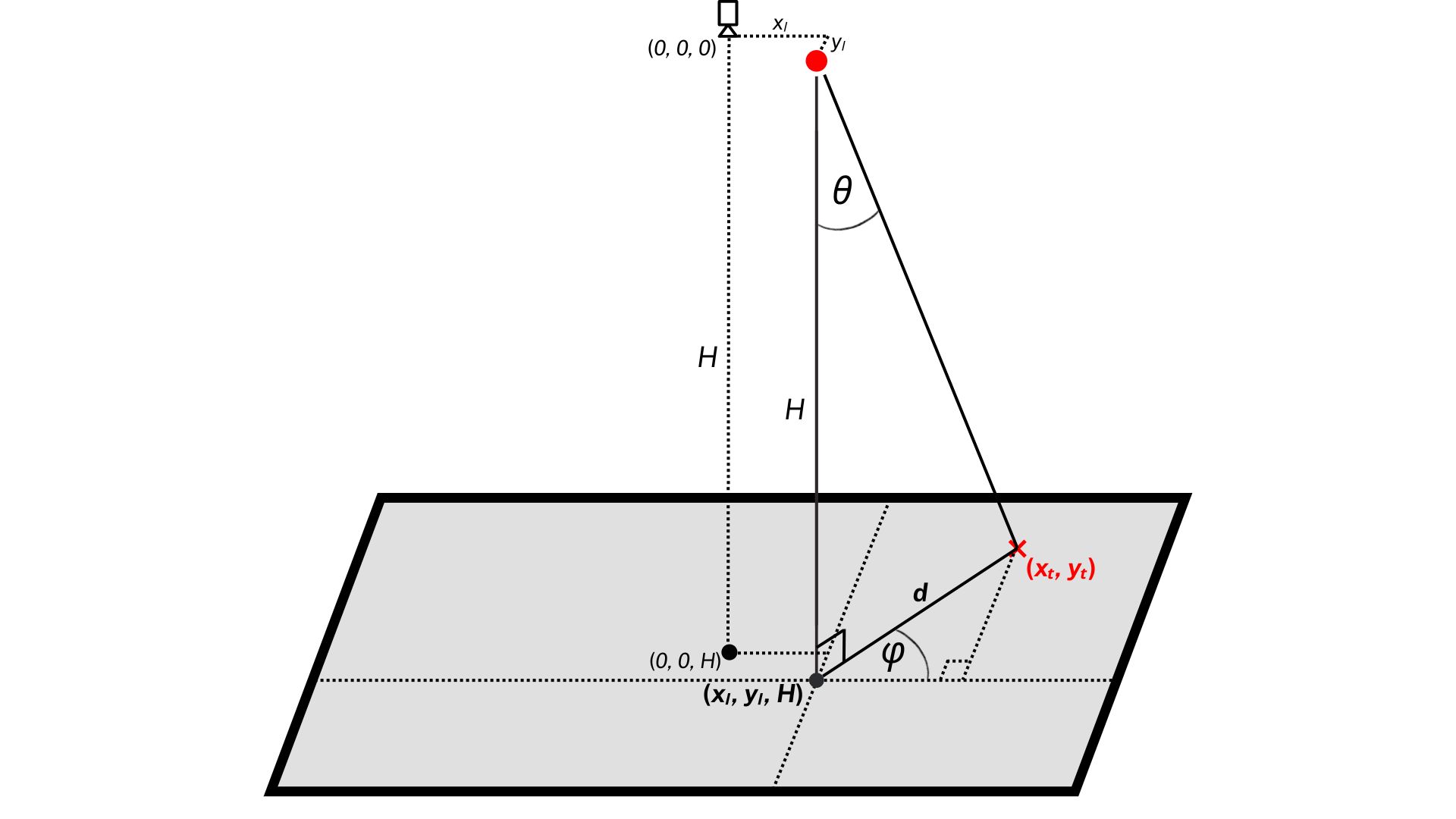}
    }
    \caption{Calculating angles $\varphi$ and $\theta$}
    \label{tracking_geometry}
\end{figure}

When performing calculations, the camera sensor is treated as the origin point of the Cartesian coordinate system (Fig.~\ref{tracking_geometry}). We treat the image taken by the camera as a 2D plane at a distance $H$ from the camera sensor. We physically measure the value of $H$ as the height of the camera sensor in relation to the input of the washing unit. The laser's origin point is located at $(x_l,y_l,z_l)$. The final prototype will be constructed in such a way that $z_l=0$, and values of $x_l$ and $y_l$ are precisely known. Also known are the coordinates of the target point on the image plane, marked as $(x_t,y_t)$. 

With these values known, we can derive the values for $\varphi$ - the pan angle, and $\theta$ - the tilt angle. We project the point of the laser origin onto the image plane. We then calculate $d$ - the distance between this projection and the target point:

$$d = \sqrt{(x_t - x_l)^2 - (y_t - y_l)^2}$$

Now we can calculate $\varphi$ and $\theta$:

$$\varphi = \arccos{(\frac{x_l + x_t}{d})}, \quad \theta = \arctan{(\frac{d}{H})}$$

These values need to be normalised to a value between $-1$ and $1$, which corresponds to the servos' rotation range of $-90^{\circ}$ to $90^{\circ}$ respectively. Depending of the quadrant which the target point occupies, the values may also need to be inverted.

This method of determining the servo angles is precise if we assume that the target is at height $H$ from the device. This, of course, will not be the case most of the time. A rock can be deep in the washing unit or high on top of a pile of olives, and there is little we can do to estimate the exact height at which it is located from the 2D image alone. We can tackle this problem by either modifying the hardware or the software of our device.

From the hardware side, we could add a second camera to obtain binocular vision, granting us the ability to estimate depth of the environment the sensors are observing. At a higher cost, we could instead mount a reputable 3D camera which would precisely calculate the distance from the lens to every visible point. This remains an option for the future, but for the prototype solution we opted to correct errors in laser tracking with software. Once the initial angles are determined and the two servos take their positions, the laser turns on and the camera takes a picture of the scene. On the picture, the laser dot can be located with little difficulty. Once the dot is detected, we can calculate the error caused by the depth variance and adjust the tilt servo accordingly. 

\section{Model evaluation and performance}

We evaluate the performance of the trained models primarily based on their Average Precision (AP)\footnote{AP is a widely used metric in computer vision projects, which combines metrics such as precision, recall and IoU (Intersection over Union) into a single value that gives a good indication of overall model performance.}. This metric alone, however, will not be enough to properly evaluate the model's performance, because in this case we value some aspects of the model more than the others. For example, we want the recall metric to be as high as possible, which measures the ratio between the number of correctly detected objects and number of occurring objects in total. The model is preferred to make some false alarms while detecting most of the rocks, rather than being precise and not raising an alert unless it is absolutely certain that it spotted a rock. Thus, while it is important to keep the AP metric as high as possible, we will also be focusing on keeping the recall metric high by tuning the model to prefer scoring well on this metric over precision.

At the time of writing this paper, our best performing model scored 85.3\% precision with 52.4\% recall on never-seen-before test images. These results suggest that our device is able to detect more than half of the appearing rocks with high certainty, raising fairly little false alarms. Upon investigation of the results, it is evident that the smaller the rock appears to be, the less likely it is to be detected. In the following months we will keep working on training new models, ensuring that the performance gets even better for the final prototype. Once we install the prototype, it will be able to gather more data during its use, ensuring that the model keeps evolving and improving as time passes.

\section{Acknowledgment}

We would like to thank Lučka Olive for collaborating with us and granting us unhindered access to their machinery. This work would not have been possible without their open-minded attitude towards new technologies and willingness to take risks in order to modernise and improve the process of olive oil production.

This work was partially supported by Erasmus+ Project No. 2022-1-PL01-KA220-HED-000088359 entitled "The Future is in Applied Artificial Intelligence" (FAAI) \cite{FAAI}, which aims to join together Higher Education Institutions (HEI) and businesses. In this context, this project has to bridge the current AI skills gap, build an AI ecosystem of key partners, promote AI business opportunities, and support the creation of internship programs in AI. The FAAI project activities focus on HEI trainers, undergraduate and postgraduate students, and business managers. Furthermore, the project is promoting among businesses and young people the enormous opportunities provided by AI to build the ecosphere of modern society. The given work was performed within the framework of the FAAI work package 4 entitled "Artificial Intelligence framework for training in HE" and presents a real use case that is offered for studying applied AI.


\begin{thebibliography}{00}

\bibitem{almonds} Son, G.J., Kwak, D.H., Park, M.K., Kim, Y.D., \& Jung, H.C. (2021). U-Net-Based Foreign Object Detection Method Using Effective Image Acquisition System: A Case of Almond and Green Onion Flake Food Process. Sustainability, 13(24).

\bibitem{dHash} David Oftedal (2014), Difference Hash - An algorithm for comparing images based on their visual characteristics (2014), \url{https://01101001.net/differencehash.php}

\bibitem{SAHI} Akyon, F., Altinuc, S., \& Temizel, A. (2022). Slicing Aided Hyper Inference and Fine-tuning for Small Object Detection. 2022 IEEE International Conference on Image Processing (ICIP), 966-970.

\bibitem{albumentations} Buslaev, Alexander \& Parinov, Alex \& Khvedchenya, Eugene \& Iglovikov, Vladimir \& Kalinin, Alexandr. (2018), Albumentations: fast and flexible image augmentations. 

\bibitem{CLAHE} Stephen M. Pizer, E. Philip Amburn, John D. Austin, Robert Cromartie, Ari Geselowitz, Trey Greer, Bart ter Haar Romeny, John B. Zimmerman, Karel Zuiderveld, Adaptive histogram equalization and its variations, \url{https://www.sciencedirect.com/science/article/pii/S0734189X8780186X}

\bibitem{YOLOv8} Jocher, G., Chaurasia, A., \& Qiu, J. (2023). Ultralytics YOLO (Version 8.0.0) [Computer software]. \url{https://github.com/ultralytics/ultralytics}

\bibitem{FAAI} The Future is in Applied Artificial Intelligence (FAAI).  (2022-2024), \url{https://faai.ath.edu.pl}

\end{thebibliography}
\end{document}